\pdfoutput=1

\documentclass[11pt]{article}

\usepackage[preprint]{acl}

\usepackage{times}
\usepackage{latexsym}

\usepackage[T1]{fontenc}

\usepackage[utf8]{inputenc}

\usepackage{microtype}

\usepackage{inconsolata}

\usepackage{graphicx}

\usepackage{microtype}
\usepackage{graphicx}
\usepackage{subfigure}
\usepackage{booktabs} 
\usepackage{tcolorbox}

\usepackage{hyperref}

\usepackage{amsmath,amsfonts,bm}
\usepackage{amssymb}
\usepackage{mathtools}
\usepackage{amsthm}

\def\vd{{\bm{d}}}

\def\sD{{\mathbb{D}}}

\usepackage[capitalize,noabbrev]{cleveref}
\newcommand{\minisection}[1]{\noindent{\textbf{#1}}.}

\theoremstyle{plain}

\theoremstyle{definition}

\theoremstyle{remark}

\title{ Enhancing LLM Knowledge Learning through Generalization}

\author{\textbf{Mingkang Zhu\textsuperscript{1}},
\textbf{Xi Chen\textsuperscript{2}},
\textbf{Zhongdao Wang\textsuperscript{3}},
\textbf{Bei Yu\textsuperscript{1}}
\textbf{Hengshuang Zhao\textsuperscript{2}}
\textbf{Jiaya Jia\textsuperscript{4,5}}
\\\\
\textsuperscript{1}CUHK
\textsuperscript{2}HKU
\textsuperscript{3}Huawei
\textsuperscript{4}SmartMore
\textsuperscript{5}HKUST\\
\texttt{mkzhu23@cse.cuhk.edu.hk, jia@cse.ust.hk}  \\
}

\begin{document}
\maketitle

\begin{abstract}

As Large language models (LLMs) are increasingly deployed in diverse applications, faithfully integrating evolving factual knowledge into these models remains a critical challenge. Continued pre-training on paraphrased data has shown empirical promise for enhancing knowledge acquisition. However, this approach is often costly and unreliable, as it relies on external models or manual effort for rewriting, and may inadvertently alter the factual content. In this work, we hypothesize and empirically show that an LLM's ability to continually predict the same factual knowledge tokens given diverse paraphrased contexts is positively correlated with its capacity to extract that knowledge via question-answering. Based on this view and aiming to improve generalization to diverse paraphrased contexts, we introduce two strategies to enhance LLMs' ability to predict the same knowledge tokens given varied contexts, thereby enhancing knowledge acquisition. First, we propose formatting-based data augmentation, which diversifies documents conveying the same knowledge by altering document formats rather than their content, thereby preserving factual integrity. Second, we adopt sharpness-aware minimization as the optimizer to better improve generalization. Extensive experiments demonstrate our methods' effectiveness in both continued pre-training and instruction tuning, and further gains can be achieved by combining with paraphrased data.

\end{abstract}

\section{Introduction}
\label{intro}

Large language models (LLMs) are pre-trained on large-scale corpora  encompassing extensive knowledge, enabling them to demonstrate remarkable performance on knowledge-intensive tasks \cite{gpt3-brown-2020,gpt4-2023,palm-chowdhery-2022,opt-zhang-2022,llama-touvron-2023,llama2-touvron-2023,gemini-2023,qwen2}. As LLMs are increasingly deployed in real-world applications, an essential challenge is the effective integration of evolving factual knowledge. Prior work has shown that continued pre-training on a single document is insufficient for reliable knowledge elicitation via question answering (QA) \cite{kandpal2023largelanguagemodelsstruggle}. In contrast, training on multiple paraphrased versions of a document has been empirically shown to improve knowledge acquisition \cite{instruction-tuned,physicslanguagemodels31}. However, this method incurs significant costs and presents reliability concerns. Specifically, it typically depends on either computationally intensive external models or manual rewriting and has the risks of inadvertently altering factual information \cite{data_aug_using_LLM}.

To address these challenges, we begin by analyzing the empirical observation that training with paraphrased documents improves QA accuracy for the corresponding embedded knowledge  \cite{physicslanguagemodels31}. We interpret diverse documents expressing the same factual content as samples drawn from a shared underlying distribution. From this perspective, training on a document containing certain knowledge tokens enhances LLMs' ability to predict corresponding knowledge tokens given diverse unseen paraphrased preceding contexts. Further improvements can be achieved by training on paraphrased versions containing the same factual content, which enhances the model's generalization to varied preceding contexts. Based on this insight,  we hypothesize a positive correlation between an LLM's ability to continually predict factual knowledge tokens given diverse unseen paraphrased contexts and its ability to extract that knowledge through QA, as demonstrated in \cref{fig:teaser}. This connection is not immediately intuitive from a human perspective, as declarative documents and question-answer formats differ significantly in structure.

To empirically validate our hypothesis, we construct a biography dataset with diverse attributes, following the methodology in \citet{physicslanguagemodels31}. Our observation reveals a strong positive correlation between the model’s accuracy in predicting knowledge tokens given unseen paraphrased contexts, and its accuracy in answering questions about the same knowledge. Additionally, integrating paraphrased documents in training enhances both accuracies, further reinforcing our hypothesis. These findings suggest that enhancing LLMs' ability to predict knowledge tokens conditioned on varied contexts is a promising direction for improving knowledge acquisition.

Motivated by the findings, we propose two strategies to make accurate knowledge token prediction generalize to diverse paraphrased preceding contexts, thereby improving knowledge acquisition. As knowledge learning from a single document represents one-shot learning, while incorporating paraphrased documents transforms it into few-shot learning, we first propose the formatting-based data augmentation to diversify the documents containing the same knowledge. This approach modifies training documents' formats—such as changes in spacing and padding—as a form of data augmentation. It eliminates the reliance on calling external models or manual labor and does not risk introducing factual inconsistencies. Second, we introduce sharpness-aware minimization (SAM) \cite{foret2021sharpnessaware} as the optimization method to better improve generalization. When treating preceding document contexts as the input and knowledge tokens as targets, SAM facilitates generalization to diverse preceding document contexts with the same knowledge tokens. Furthermore, recent works identify that including different question paraphrases in instruction tuning helps knowledge extraction \cite{fu-etal-2024-learning}. Thus, we additionally incorporate our method into instruction tuning to improve the model's generalization on question paraphrases.

We evaluate our methods on our constructed biography dataset following \citet{physicslanguagemodels31}, and the Wiki2023 dataset \cite{instruction-tuned}. Experiment results demonstrate significant improvements in LLMs' knowledge learning ability for both the continued pre-training and instruction tuning phases. Moreover, combining our  methods with paraphrased data leads to more improvement. Our key contributions can be summarized as follows:

\begin{itemize}
\item We hypothesize and empirically demonstrate that an LLM's ability to continue knowledge tokens given diverse paraphrased contexts is positively correlated with its capacity to answer related questions. This analysis provides a novel perspective on enhancing knowledge acquisition in LLMs and motivates us to improve the ability from the generalization perspective.

\item We propose formatting-based data augmentation to automatically generate diverse training documents without relying on external paraphrasing tools or introducing factual inconsistencies. In addition, we propose to employ SAM as an optimization technique to boost generalization. We further apply our method on instruction tuning, enabling better generalization across paraphrased questions and enhancing knowledge extraction.

\item Extensive experiments and ablation studies demonstrate our method's effectiveness in improving factual knowledge learning during both the continued pre-training and instruction tuning phases.
\end{itemize}

\section{Related Work}
\label{related}

\minisection{Continued LLM Knowledge Learning}
As the pre-trained knowledge stored in LLMs quickly becomes outdated, adapting up-to-date information into LLMs becomes a critical problem. The primary approach to tackle this problem is through continued pre-training on documents containing up-to-date knowledge \cite{ovadia2024finetuningretrievalcomparingknowledge, instruction-tuned, jang2022towards}. However, straightforward training on new corpus usually cannot lead to effective knowledge acquisition. This is likely due to the lack of diverse knowledge demonstrations like foundational or textbook knowledge \cite{physicslanguagemodels31,instruction-tuned, ovadia2024finetuningretrievalcomparingknowledge, cheng2024adapting}. Therefore, some works focus on paraphrasing documents to alleviate this issue \cite{cheng2024adapting, physicslanguagemodels31,ovadia2024finetuningretrievalcomparingknowledge}. However, paraphrasing manually or using external models can be expensive and tedious, and it might be unreliable as facts and knowledge inside documents could be altered \cite{data_aug_using_LLM}. Therefore, we aim to avoid the risk of changing facts embedded in documents while enabling effective knowledge acquisition. Some works try to include QA data together with or before adapting to new documents \cite{physicslanguagemodels31, instruction-tuned}. However, these methods introduce new difficulties in finding effective arrangements and proportions of QA data and documents. To induce effective knowledge extraction during inference, instruction tuning on annotated QA pairs after continued pre-training on documents has become a common practice \cite{t0-sanh-2022,flan-jason-2022,crosstask-2022-mishra,iyer-optiml-2022,openassistant-kopf-2023,lima-zhou-2023, salmon-2023-sun,principle-zhiqing-2023, fu-etal-2024-learning}. Thus, we also aim to achieve satisfactory knowledge extraction after instruction tuning.

\minisection{Understanding of LLM Knowledge Learning}
Several works are trying to understand how LLMs learn knowledge from documents and retrieve them in question answering. \citet{akyurek2022tracing} tries to detect training documents important for question-answering for pre-trained LLMs. A number of works find connections between the frequency of certain knowledge appearing in pre-training documents and its question-answering ability \cite{kandpal2023largelanguagemodelsstruggle, akyurek2022tracing, lama-petroni-2019, kassner2020pretrained, wei2021frequency, fevry2020entities, de2020autoregressive}. Recently, \citet{physicslanguagemodels31} and \citet{ovadia2024finetuningretrievalcomparingknowledge} empirically observe that adding paraphrased documents in the pre-training and continued pre-training phase helps knowledge extraction.

\minisection{Data Augmentation for Natural Language Processing}
There is a rich literature on data augmentation techniques in natural language processing \cite{chen2021empiricalsurveydataaugmentation, data_aug_using_LLM, wei2019eda}. A popular type of data augmentation is synonym substitution, which replaces words in documents with their synonyms according to pre-defined dictionaries \cite{kolomiyets-etal-2011-model, Yang2015That, Zhang2015Character}. Another popular class of data augmentation is inserting, replacing, deleting, and swapping words in documents \cite{wei2019eda, iyer-optiml-2022,niu2018adversarial, miao2020snippext}. In the era of LLMs, paraphrasing documents or synthesizing data using LLMs become increasingly popular \cite{data_aug_using_LLM, sharma-etal-2023-team, nair-etal-2023-generating}. However, these data augmentation methods generally modify the semantics of original documents, and certain tokens in documents where the factual knowledge resides have the risk of being altered \cite{data_aug_using_LLM}. Therefore, we opt to avoid such risks and design suitable data augmentation methods for LLM knowledge learning.

\section{Analysis of LLM Knowledge Learning}
\label{sec1}
In this section, we analyze the practical effectiveness of paraphrasing for knowledge learning, aiming to derive insights that inform the development of more effective knowledge acquisition methods.

\subsection{Autoregressive Language Model} \label{subsec3.1}
Let $\boldsymbol{\theta}$ denote the parameters of an autoregressive language model, and $\mathcal{V}$ be a fixed vocabulary of tokens. Suppose document $\vd$ contains a sequence of tokens $(x_1, x_2, \dots, x_T)$ from $\mathcal{V}$, where $T$ is the length of the document. The training sequence has a special token $x_0 = $ $<$bos$>$ prepended, indicating the sequence's start. The autoregressive language modeling task estimates the conditional probability distribution $P(x_t|x_{<t})$ for each $t = 1, 2, \dots, T$. This is typically achieved by training a deep neural network to predict the next token $x_t$ given the previous tokens $x_0, x_1, x_2, \dots, x_{t-1}$. For document $\vd$, the negative log-likelihood loss function of the observed sequence $\vd$ is:
\begin{equation}\label{loss} 
\begin{aligned}
\ell(\boldsymbol{\theta}, \vd)  & = -\log P_{\boldsymbol{\theta}}(\vd)  \\
& = -\log P_{\theta}(x_1, x_2, \dots, x_T \mid x_0)   \\
   & = -\log\prod_{t=1}^{T} P_{\boldsymbol{\theta}}(x_t|x_{<t}) \\
  & =  -\sum_{t=1}^{T} \log P_{\boldsymbol{\theta}}(x_t|x_{<t}).
\end{aligned}          
\end{equation}

\subsection{Connecting Paraphrasing with the Effectiveness of Knowledge Learning} \label{subsec3.2}

\begin{figure}[t]
  \centering  
\includegraphics[width=0.48\textwidth]{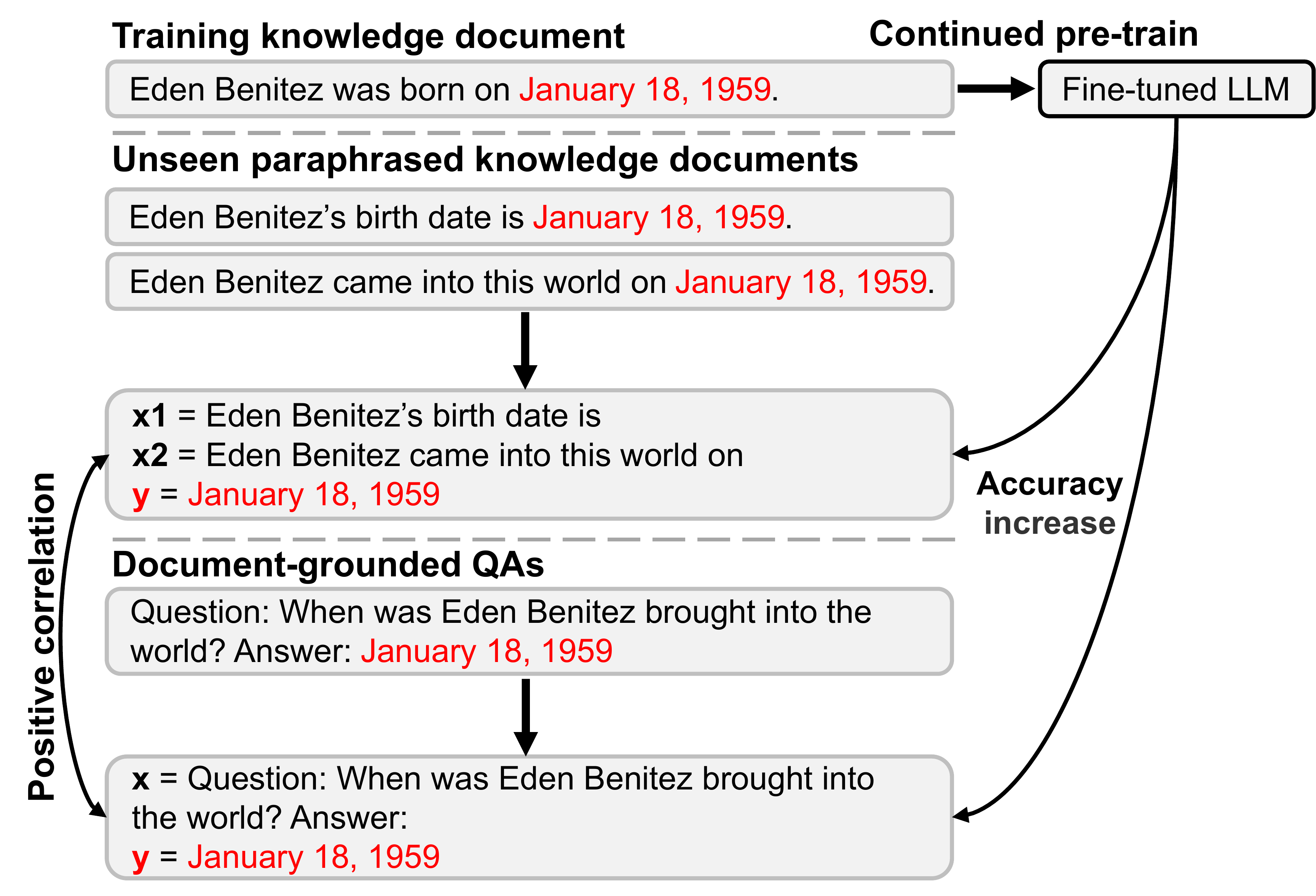 }
  \caption{Conceptual illustration of our hypothesis. The red tokens represent knowledge tokens. Continued pre-training on raw knowledge documents leads to prediction accuracy increase of knowledge tokens conditioned on both preceding contexts of unseen paraphrased documents containing the same knowledge and related questions. Incorporating paraphrased training documents further improves both accuracies. Thus, their accuracies are positively correlated.}
  \label{fig:teaser}
\end{figure}

\begin{figure*}[t]
  \centering  
\includegraphics[width=0.97\textwidth]{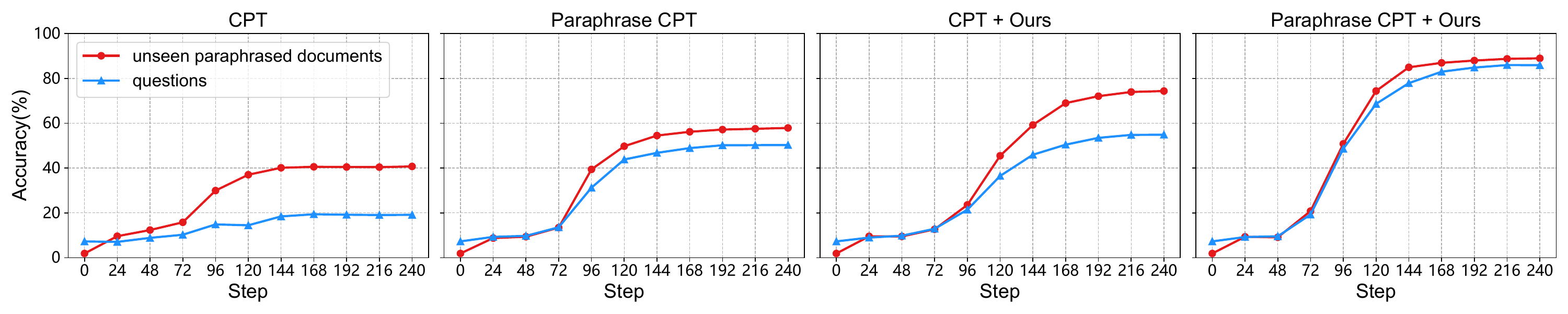 }
  \vspace{-0.2cm}
  \caption{Average first knowledge token accuracy for each attribute conditioned on: (1) preceding tokens in unseen paraphrased documents, (2) preceding tokens in testing questions.}
  \label{fig:first_token}
\end{figure*}

Building on prior empirical findings that training with paraphrased documents containing the same knowledge enhances knowledge learning \cite{instruction-tuned,physicslanguagemodels31}, we investigate the underlying mechanism driving this effect. We conceptualize diverse documents that convey the same knowledge as samples drawn from a shared distribution. From this perspective, training on documents containing specific knowledge tokens improves an LLM’s ability to predict those knowledge tokens given diverse unseen paraphrased preceding contexts. This ability is further strengthened by training on paraphrased variants, which promotes generalization across varied preceding context expressions. Therefore, we hypothesize a positive correlation between an LLM's ability to continue factual knowledge tokens from diverse paraphrased contexts and its ability to extract that knowledge through QA. The conceptual illustration of our hypothesis is demonstrated in \cref{fig:teaser}. However, this is not obvious from the human perspective, as documents and questions asking for embedded knowledge are different: one is a declarative sentence, while the other is in the form of QA. Therefore, we need to verify this hypothesis.

\subsection{Verification} \label{subsec3.3}
To validate our hypothesis in a controlled setting, we generate synthetic human biography data following the methodology of \citet{physicslanguagemodels31}. Specifically, we create 1,000 randomly generated human biography profiles, each characterized by five attributes: birth date, college, major, hometown, and company. To construct training documents, we employ a predefined template to represent these five attributes and leverage ChatGPT to generate two additional paraphrased templates. Each profile's attributes are populated into these templates to form the training documents. For evaluation, we generate five additional paraphrased biography documents and create five questions for each attribute. The templates, example documents, and QAs are included in Appendix.

Then we conduct a study on two scenarios: (1) \textbf{CPT}, each biography profile has a single document representation; (2) \textbf{Paraphrase CPT}, each biography profile has 3 paraphrased document representations. We continually pre-train Qwen 2 1.5B \cite{qwen2}  with the above settings and record the first knowledge token's accuracy conditioned on preceding contexts tokens in 5 unseen paraphrased documents and related testing questions. If the model's accuracy predicting the first knowledge tokens conditioned on preceding tokens in unseen documents increases along with that conditioned on related questions, a positive correlation exists between these two metrics. The results are presented in \cref{fig:first_token}.

\begin{table}[t]
\centering
\scalebox{0.63}{
\begin{tabular}{c c c c c}
\toprule
\textbf{Setting } & \textbf{CPT} & \textbf{Para. CPT} & \textbf{CPT  + Ours} & \textbf{Para. CPT + Ours}\\
 \midrule
Pearson corr. & 0.97 & 1.00 &1.00&1.00\\
Spearman corr.  & 0.95&  1.00 & 0.99&0.98\\ 
 \bottomrule
\end{tabular}}
\caption{
Pearson and Spearman correlation between the prediction accuracy of first knowledge tokens for each attribute conditioned on: (1) preceding tokens in unseen paraphrased documents, (2) testing questions.}
\label{tbl:correlation}
\end{table}

In \cref{fig:first_token}, red lines represent the prediction accuracy of first knowledge tokens conditioned on their preceding tokens in unseen paraphrased documents, while blue lines represent the accuracy of first knowledge tokens conditioned on testing questions. We can see that the accuracy conditioned on questions is increasing along with the accuracy conditioned on preceding tokens in unseen paraphrased documents. Moreover,  this also holds true for training on paraphrased documents, which leads to much higher accuracies. We further calculate the Pearson and Spearman correlation between the accuracies given unseen documents and questions in \cref{tbl:correlation}, which also demonstrate a strong positive correlation between these two metrics. These observations validate the positive correlation between the the LLM’s ability to continue factual knowledge tokens given diverse unseen paraphrased contexts and its ability to extract that knowledge through QA. Hence,  enhancing LLMs’ ability to predict knowledge tokens conditioned on varied document contexts is a promising direction for improving knowledge acquisition.

\section{Methods} \label{method}

\cref{sec1} has revealed that an LLM's ability to continue  knowledge tokens given diverse paraphrased contexts is strongly positively correlated with its capacity to answer related questions. Thus, in this section, we explore two methods to improve LLM knowledge learning, by enabling accurate prediction of knowledge tokens to generalize from training documents to unseen paraphrased variants that convey the same factual content.

\subsection{Formatting-Based Data Augmentation}
Based on the analysis in \cref{subsec3.3}, knowledge learning is one-shot without paraphrasing and few-shot with paraphrased documents. Thus, providing diverse documents containing the same knowledge is crucial for effective knowledge learning. However, paraphrasing manually  or via external models can be expensive and laborious, and might alter the embedded facts \cite{data_aug_using_LLM}. Moreover, certain expressions and terminologies are irreplaceable and must be used in their exact form. Nor should mottoes and poems be paraphrased when they are in training documents. Therefore, we aim to develop methods to reliably increase the variety of training documents containing the same knowledge  without paraphrasing. 

We draw inspiration from the formatting of texts. We may often encounter variations in the formatting used to present texts, such as whether to indent the beginning of a paragraph and whether to use spaces or tabs as indentations for codes. There are also variations for using single-space or double-space spacing in the era of typewriters \cite{wiki_space}. These formatting differences, while altering some of the format tokens, do not affect the semantic meaning and knowledge of the text itself. Therefore, given a training document, we propose to apply the following formatting-based data augmentations:
\begin{itemize}
  \item \textbf{Wrapping.} Augmented documents are created by wrapping the document with quotes, asterisks, brackets, or parentheses. This is used to mimic the case that the document is quoted, highlighted, or appears in Markdown.
  \item  \textbf{Left Padding.}  Augmented documents are created by padding spaces, tabs, or pound signs to the left of the document. This is to mimic the scenarios of the document being written using Markdown or appearing as a paragraph in a paper.
  \item \textbf{Random Space Insertion.} Augmented documents are created by randomly inserting additional spaces adjacent to the original spaces. This simulates the case that the training document is presented using different spacing and includes some unintentional extra spaces.
\end{itemize}

Some augmentation examples are as follows:

\begin{tcolorbox}[width=0.48\textwidth,
    colback=gray!10!white,     
    colframe=black,            
    boxrule=1pt,               
]
{\scriptsize
\textbf{Raw document}
\begin{verbatim}
Eden Benitez was raised in Santa Clarita.
\end{verbatim}
\textbf{Wrapping Augmentation}
\begin{verbatim}
*Eden Benitez was raised in Santa Clarita.*
\end{verbatim}
\textbf{Left Padding Augmentation}
\begin{verbatim}
# Eden Benitez was raised in Santa Clarita.
\end{verbatim}
\textbf{Random Space Insertion Augmentation}
\begin{verbatim}
Eden  Benitez was raised  in Santa  Clarita.
\end{verbatim}
}
\end{tcolorbox}

Detailed specifications of the data augmentations are discussed in Appendix. With these augmented documents, we automatically diversify the training documents while not changing the knowledge and facts inside these documents.

\subsection{Generalization Regularization}

Given a training document $\vd$, the autoregressive objective in \cref{loss} would minimize the negative log-likelihood (NLL) loss for a set of samples $(x_{<t}, x_t)$. Thus, we can further enhance the generalization ability by applying generalization regularization methods designed for traditional supervised problems. Applying generalization regularization on samples with $x_t$ being knowledge tokens and $x_{<t}$ being the corresponding preceding tokens in documents can generalize to samples with the same knowledge token label $x_t$,  but different preceding tokens, such as $x_{<t}$ in unseen paraphrased documents. Recently,  \citet{foret2021sharpnessaware} developed the {\it Sharpness-Aware Minimization} (SAM) to improve the generalization ability for supervised problems\cite{baek2024why, chen2022when, foret2021sharpnessaware}. We adopt this technique to further improve LLM knowledge learning.

Given a training document $\vd$, let $\mathcal{B}$ be the set of training document samples with knowledge tokens as pseudo-label and preceding tokens as inputs, and according to SAM we solve the following problem:
\begin{align}\label{SAM}
    \min_{\boldsymbol{\theta}} ~ \max_{\|\boldsymbol{\epsilon}\|_2\le \rho} L_{\mathcal{B}}(\boldsymbol{\theta}+\boldsymbol{\epsilon}) + \lambda \|\boldsymbol{\theta}\|_2^2, 
 \end{align}
where  $\rho\ge 0$ is a given perturbation radius,  
$\lambda$ is a small positive regularization constant. The objective is to find a minimizer with the neighborhood where the loss does not increase too much. According to SAM, the inner maximization problem in \cref{SAM} 
is solved approximately at $\boldsymbol{\hat\epsilon}= \rho {\nabla L_{\mathcal{B}}(\boldsymbol{\theta})}/{\| \nabla L_{\mathcal{B}}(\boldsymbol{\theta})\|_2}$ by the first-order Taylor expansion. Then, the objective function of  \cref{SAM} changes to $L_{\mathcal{B}}(\boldsymbol{\theta}+\boldsymbol{\hat\epsilon}) + \lambda \|\boldsymbol{\theta}\|_2^2$, on which the gradient descent is performed. 

As shown in \cref{fig:first_token} and \cref{tbl:correlation}, our hypothesis in \cref{subsec3.2} continues to hold with our proposed methods, further supporting its empirical validity.

\subsection{Adaptation to Instruction Tuning}
\label{sec:instruction}
Instruction tuning has become a common practice to make LLMs follow human instructions and perform question-answering \cite{t0-sanh-2022,flan-jason-2022,instructgpt3-ouyang-2022}. Instruction tuning computes the negative log-likelihood loss only on tokens in answers with questions as the context: $L_{\bm{a}} = -\sum_{t}{\log P(\bm{a}_t|\bm{q}, \bm{a}_{<t})}$. Recent works identify that including diverse question paraphrases in instruction tuning helps knowledge extraction \cite{fu-etal-2024-learning}. Therefore, we propose to use SAM and apply our formatting-based data augmentation to the questions for instruction tuning. In this way, LLM would be able to respond accurately to different paraphrases of a question during instruction tuning, thereby enhancing knowledge extraction. 

\section{Experiments}

\subsection{Experiment Settings}

\minisection{Baseline Methods}
We experiment with two standard baselines: (1) continued pre-training (CPT) and (2) continued pre-training with instruction tuning (IT), and demonstrate the effectiveness of our methods in improving the knowledge learning abilities of these baselines.

\minisection{Base Models}
We use Qwen 2 1.5B \cite{qwen2} and LLaMA 2 7B \cite{llama2-touvron-2023} as base models, and test all baselines and their combination with our methods on these models.

\minisection{Datasets} 
We use our biography dataset  generated following \citet{physicslanguagemodels31}, and the Wiki2023-film dataset \cite{instruction-tuned} for the experiment. For the biography dataset, we follow \citet{physicslanguagemodels31} to continually pre-train on documents of all biography profiles and instruction-tune on 1 QA pair per attribute for half of the biography profiles. The evaluation is conducted for the remaining half individuals. Our evaluation differs from \citet{physicslanguagemodels31}, which evaluates only 1 question prompt per attribute. We generate 5 different question prompts for each attribute to better evaluate the generalization ability, totaling 12500 QA pairs. The biography dataset is synthetic while the recipe for generating the Wiki2023-film dataset minimizes overlap with the pre-training corpus. Thus,  experimenting on these two datasets can mimic the difficult case of continued knowledge learning on up-to-date information.

For the biography dataset, we experiment with both the \textbf{CPT} and \textbf{Paraphrase CPT} settings illustrated in \cref{subsec3.3}. For the Wiki2023-film dataset, we experiment with the \textbf{CPT} setting since there are no paraphrased documents. All comparing methods are trained with the \textbf{same number of steps} in both the continued pre-training and instruction tuning phases for fair comparison.

\minisection{Evaluation Metrics}
As we aim to evaluate the closed-book free-form question-answering ability, we utilize exact match (EM) between the model generations and ground truth answers as the evaluation metric \cite{kwiatkowski-etal-2019-natural}. We also report Recall and F1 scores to better assess questions with long answers. When evaluating models that have not been instruction-tuned, we prepend 1 QA pair for the biography dataset and 5 QA pairs for the Wiki2023-film dataset to make sure that models can follow the QA format.

\subsection{Main Results}

\begin{table*}[t]
    \centering
   \scalebox{0.8}{ 
    \begin{tabular}{@{}lccc|ccc|ccc|ccc@{}}
    \toprule
    & \multicolumn{3}{c|}{\textbf{Qwen 2 1.5B}} & \multicolumn{3}{c|}{\textbf{Qwen 2 1.5B w/ IT}} & \multicolumn{3}{c|}{\textbf{LLaMA 2 7B}} & \multicolumn{3}{c}{\textbf{LLaMA 2 7B w/ IT}} \\
    \cmidrule(l){2-4} \cmidrule(l){5-7} \cmidrule(l){8-10} \cmidrule(l){11-13}
    & \textbf{EM} & \textbf{Recall} & \textbf{F1} & \textbf{EM} & \textbf{Recall} & \textbf{F1} & \textbf{EM} & \textbf{Recall} & \textbf{F1} & \textbf{EM} & \textbf{Recall} & \textbf{F1}   \\ 
       \cmidrule(l){2-4} \cmidrule(l){5-7} \cmidrule(l){8-10} \cmidrule(l){11-13}
  \textbf{Base} &0.7 & 7.1& 6.2 & - & - & -  &0.7& 8.9 & 7.6 & - & -  & - \\
  \midrule
  \textbf{CPT} & 7.1 & 16.1& 12.4 & 52.8&57.6&57.1  & 52.0& 59.1 & 58.5 & 89.6 & 91.3  & 91.2 \\
  \textbf{w/ Ours} & \textbf{43.2} &\textbf{57.7}& \textbf{52.3} & \textbf{57.9} &\textbf{62.2}& \textbf{61.8}  &\textbf{85.4}& \textbf{88.2} & \textbf{88.0} & \textbf{93.3}& \textbf{94.2} & \textbf{94.1} \\
  \midrule
  \textbf{Paraphrase CPT} & 24.9 & 45.4& 35.1 &54.5& 59.9 &59.6& 54.3& 70.6 & 63.6 & 94.2 & 95.9  & 95.9 \\
  \textbf{w/ Ours} & \textbf{74.9} & \textbf{80.4}& \textbf{80.0} & \textbf{75.3}& \textbf{77.2} & \textbf{76.9} & \textbf{89.2}& \textbf{93.1}& \textbf{92.8} & \textbf{98.4} & \textbf{99.0} &\textbf{ 99.0}\\
    \bottomrule
    \end{tabular}}
        \caption{Experiment results on the biography dataset with the base models continued pre-training and instruction-tuning by our method and baselines. Our method leads to substantial improvement in knowledge acquisition and extraction for both phases compared to baselines.}    
    \label{tab:biography}
    \end{table*}

\begin{table*}[t]
    \centering
    \scalebox{0.83}{
    \begin{tabular}{@{}lccc|ccc|ccc|ccc@{}}
    \toprule
    & \multicolumn{3}{c|}{\textbf{Qwen 2 1.5B}} & \multicolumn{3}{c|}{\textbf{Qwen 2 1.5B w/ IT}} & \multicolumn{3}{c|}{\textbf{LLaMA 2 7B}} & \multicolumn{3}{c}{\textbf{LLaMA 2 7B w/ IT}} \\
    \cmidrule(l){2-4} \cmidrule(l){5-7} \cmidrule(l){8-10} \cmidrule(l){11-13}
    & \textbf{EM} & \textbf{Recall} & \textbf{F1} & \textbf{EM} & \textbf{Recall} & \textbf{F1} & \textbf{EM} & \textbf{Recall} & \textbf{F1} & \textbf{EM} & \textbf{Recall} & \textbf{F1}   \\ 
       \cmidrule(l){2-4} \cmidrule(l){5-7} \cmidrule(l){8-10} \cmidrule(l){11-13}
  \textbf{Base} &3.4 & 7.2& 7.5 & - & - & -  &5.6& 18.8 & 16.9 & - & -  & - \\
  \midrule
  \textbf{ CPT} & 7.2 & 19.4& 17.8 & 12.3&23.9&24.0  & 11.8& 32.7 & 27.1 &31.3 & 47.4  &46.9 \\
  \textbf{w/ Ours} & \textbf{9.8} &\textbf{24.9}& \textbf{22.0} & \textbf{14.8} &\textbf{27.0}& \textbf{27.3}  &\textbf{17.6}& \textbf{42.3} & \textbf{34.7} & \textbf{38.6}& \textbf{56.0} & \textbf{55.2} \\
    \bottomrule
    \end{tabular}}
        \caption{Experiment results on the Wiki2023-film dataset with the base models continued pre-training and instruction-tuning by our method and baselines. Our method leads to nontrivial improvement in knowledge acquisition and extraction for both phases compared to baselines.}    
    \label{tab:wiki}
    \end{table*}

\minisection{Results on Biography Dataset}
We present the results on the biography dataset in \cref{tab:biography}.  We can see that before training, base models cannot answer questions at all. This effectively simulates the case of LLMs adapting to up-to-date information, which is considered nontrivial \cite{instruction-tuned,ovadia2024finetuningretrievalcomparingknowledge}. 
We can see that our method leads to substantial improvement over baselines for both cases of training on with and without paraphrased documents.  Notably, we observe gains of up to 50 in EM, 35 in Recall, and 44.9 in F1.

Besides, our method without paraphrased documents even significantly outperforms naively training on paraphrased documents in the continued pre-training phase, and leads to on-par performance for the instruction tuning phase. This result shows that our method can serve as an effective and reliable alternative to tedious and unreliable paraphrasing. When applying our methods to paraphrased documents, the knowledge learning performance becomes even better, showing that our methods can induce more generalization ability with more diverse documents conveying the same knowledge. This also demonstrates that our method can gain more enhancement when used together with paraphrasing. Moreover, our method leads to much better performance than baselines prior to the instruction tuning stage. The ability to extract learned knowledge at this early stage further demonstrates the effectiveness of our method in knowledge learning. This property could be beneficial in scenarios with limited resources, such as adapting LLMs to new domains where it is challenging and labor-intensive to annotate instruction-following examples.

\minisection{Results on Wiki2023 Dataset}
Next, we evaluate our method with baselines on the Wiki2023-film dataset. As it does not have paraphrased training documents, we continually pre-train using a single document for all comparing methods. As shown in \cref{tab:wiki}, our method consistently outperforms the baselines, demonstrating stable improvements in both the continued pre-training and instruction tuning phases. Notably, we observe gains of up to 7.3 in EM, 9.6 in Recall, and 8.3 in F1.

Additionally, we can observe from \cref{tab:biography} and \cref{tab:wiki} that, our approach is consistently effective across different models, training phases, and datasets, demonstrating the robustness and effectiveness of our approach. We also want to stress that all comparing methods are trained with \textbf{the same number of steps} in both the continued pre-training and instruction tuning phases. The performance gain of our approach and paraphrasing is not attributable to an increased number of training steps on enlarged datasets. This result further stresses the importance of our findings.

\subsection{Analysis}
In this section, we conduct comprehensive ablation studies on the effect of each component of our methods on both continued pre-training and instruction tuning. We also analyze the effect of our formatting-based data augmentation compared to traditional NLP augmentations.

\minisection{Effect of Our Methods on Continued Pre-training}
We first ablate the effect of our formatting-based data augmentation and SAM for the continued pre-training phase. We use Qwen 2 1.5B \cite{qwen2} as the base model and conduct experiments on our synthesized biography dataset. We can see from \cref{tbl:ablation_cp} that, when training with paraphrased documents, both SAM and formatting-based data augmentation alone can bring measurable enhancement over the baseline. Furthermore, when SAM and our data augmentation are combined, the performance gains are further amplified. When training without paraphrased documents, the lack of diverse documents conveying the same knowledge decreases the performance gain from SAM alone. Our data augmentation, on the other hand, brings adequate document variety, which leads to substantial improvement over the baseline. With the document diversity, SAM is able to boost the performance even further.

\begin{table}[t]
\centering
\scalebox{0.66}{
\begin{tabular}{c ccc ccc}
\toprule
\textbf{Training Setting} & \multicolumn{3}{c}{\textbf{ CPT}} & \multicolumn{3}{c}{\textbf{Paraphrase CPT}} \\
\cmidrule(lr){2-4} \cmidrule(lr){5-7}
 & \textbf{EM} & \textbf{Recall} & \textbf{F1} & \textbf{EM} & \textbf{Recall} & \textbf{F1}\\ \midrule
CPT  & 7.1   & 16.1    & 12.4  & 24.9         & 45.4 & 35.1 \\ 
 \midrule
w/ Format Aug.   & 24.3   & 38.6    & 33.1  & 54.9         & 78.8 & 65.3 \\ 
w/ SAM   & 19.7   & 29.1   & 26.6 & 52.8         & 63.3 & 60.8 \\ 
w/ Format Aug.+SAM  & \textbf{43.2}  & \textbf{57.7}    & \textbf{52.3} & \textbf{74.9}         & \textbf{80.4} & \textbf{80.0} \\ \bottomrule
\end{tabular}}
\caption{
Ablation study on the effect of integrating each component of our method into the continued pre-training phase.}
\label{tbl:ablation_cp}
\end{table}

\begin{table}[t]
\centering
\scalebox{0.66}{
\begin{tabular}{c ccc ccc}
\toprule
\textbf{Training Setting} & \multicolumn{3}{c}{\textbf{ CPT}} & \multicolumn{3}{c}{\textbf{Paraphrase CPT}} \\
\cmidrule(lr){2-4} \cmidrule(lr){5-7}
 & \textbf{EM} & \textbf{Recall} & \textbf{F1} & \textbf{EM} & \textbf{Recall} & \textbf{F1}\\ \midrule
IT  & 52.8   & 57.6   & 57.1  & 70.4         & 72.8 & 72.6 \\ 
 \midrule
w/ Format Aug.   & 55.3   & 59.5    & 59.2  & 73.2         & 75.1 & 74.9 \\ 
w/ SAM   & 56.0   & 60.5    & 60.0 & 73.6         & 75.8 & 75.5 \\ 
w/ Format Aug.+SAM  & \textbf{57.9}   & \textbf{62.2}    & \textbf{61.8}  & \textbf{75.3}       & \textbf{77.2} & \textbf{76.9} \\ \bottomrule
\end{tabular}}
\caption{
Ablation study on the effect of integrating each component of our method into instruction tuning.}
\label{tbl:ablation_ins}
\end{table}

\minisection{Effect of Our Methods on Instruction Tuning}
Next, we ablate the effect of our method on the instruction tuning phase. Still, we conduct experiments on our generated biography dataset. We use Qwen 2 1.5B \cite{qwen2} continually pre-trained by our method as the starting point for instruction tuning. Prior works generally consider that knowledge is learned during continued pre-training and then made extractable in the instruction tuning phase \cite{physicslanguagemodels31, instructgpt3-ouyang-2022,t0-sanh-2022,flan-jason-2022}. Therefore, starting from the same continually pre-trained model, we can analyze how our methods influence knowledge extraction in this ablation. From  \cref{tbl:ablation_ins}, we can see that both SAM and formatting-based data augmentation alone can improve knowledge elicitation over baseline instruction tuning. Furthermore, the combination of them leads to better performance, with up to 5.1 increases on EM, 4.6 increases on Recall, and 4.7 increases on F1, over baseline instruction tuning. This result shows that our method can lead to generalization on various question paraphrases, which prior works identify as important for knowledge extraction \cite{fu-etal-2024-learning}.

\minisection{Comparison with Traditional NLP Augmentation}
We compare our formatting-based data augmentation with a representative traditional NLP data augmentation technique, EDA \cite{wei2019eda}. The experiment is conducted by continually pre-training Qwen 2 1.5B on the biography dataset under the \textbf{CPT} setting. From \cref{tbl:compare_nlp_aug}, we can see that EDA is harmful to knowledge learning. The exact match decreases from 7.1 to 0 when applying EDA. EDA uses random word insertion, random word deletion, and random word swap, which might be appropriate for improving the language modeling ability. However, EDA is highly likely to alter the knowledge in documents, making it unsuitable for the knowledge learning task. This further demonstrates our formatting-based augmentation's advantage that it reliably increases document diversity without changing the embedded knowledge.

\begin{table}[t]
\centering
\scalebox{0.7}{
\begin{tabular}{c c c c}
\toprule
\textbf{Training Setting } & \textbf{EM} & \textbf{Recall} & \multicolumn{1}{c}{\textbf{F1}}\\
 \midrule
CPT & 7.1 & 16.1 &12.4\\
 \midrule
w/ EDA  & 0.0&  9.0 & 3.8\\ 
w/ Format Aug.  & \textbf{24.3}&  \textbf{38.6} & \textbf{33.1}\\ 
 \bottomrule
\end{tabular}}
\caption{
Comparison between our formatting-based data augmentation and EDA \cite{wei2019eda}.}
\label{tbl:compare_nlp_aug}
\end{table}

\section{Conclusion}

In this paper, we try to improve LLM knowledge learning without costly and unreliable paraphrasing. We hypothesize and empirically show the positive correlation between an LLM's ability to continue factual knowledge tokens given diverse paraphrased contexts and its capacity to extract that knowledge via question-answering. Based on this insight  and from the generalization perspective, we propose formatting-based data augmentation to diversify the training documents without expensive and unreliable paraphrasing. We also propose to adopt SAM to further enhance generalization. Additionally, we incorporate our methods to instruction tuning to improve knowledge extraction. Extensive experiments demonstrate our methods' effectiveness in improving knowledge acquisition and extraction for both continued pre-training and instruction tuning phases. We hope our work can provide insights to better understand and develop effective methods for LLM knowledge learning.

\section*{Limitations}\label{limitations}

This paper introduces novel methods to enhance knowledge learning in LLMs, grounded in our empirically supported hypothesis that an LLM’s ability to continue factual knowledge tokens given diverse paraphrased contexts is positively correlated with its capacity to extract that knowledge through question-answering. While our experiments consistently validate this correlation, we acknowledge that our findings are empirical in nature, and a formal theoretical justification is not provided. We view this as a natural direction for future work. While our method improves knowledge learning for LLMs, it may also amplify factual inaccuracies if the training data contains misinformation.


\end{document}